\documentclass[conference, letterpaper]{ieeeconf}
\IEEEoverridecommandlockouts   
\usepackage{hyperref}
\usepackage{tabularx}
\usepackage{ifthen}
\usepackage[T1]{fontenc}
\usepackage[utf8]{inputenc}
\usepackage[english]{babel}
\usepackage{amsmath}
\usepackage{bbm}
\usepackage{graphicx}
\usepackage[rgb,dvipsnames]{xcolor}
\usepackage{booktabs}
\usepackage{subcaption}
\usepackage{amsfonts}

\usepackage{multirow}
\usepackage{lipsum}
\usepackage{amsmath}
\usepackage{amssymb}
\usepackage[ruled,vlined]{algorithm2e}
\usepackage{graphicx}
\graphicspath{{./Figures/}}

\title{\LARGE \bf "Dear LLaVA, Please Drive": A Depth-Aware Vision-Language Agent for Closed-Loop Robotic Control}
\begin{document}
\include{bib_short.def}


\author{
Sebastian Berger$^{1}$,
Katharina Winter$^{1}$,
Fabian B. Flohr$^{1}$%
\thanks{$^{1}$Munich University of Applied Sciences, Intelligent Vehicles Lab (IVL), 80335 Munich, Germany
\texttt{intelligent-vehicles@hm.edu}}
}

\maketitle

\begin{abstract}
Vision–language models (VLMs) provide a compelling foundation for reasoning-driven mobile navigation, offering rich contextual understanding and strong generalization from large-scale pretraining. Most existing navigation frameworks rely on imitation learning and therefore require substantial labeled trajectory data, limiting their scalability and robustness. In this work, we propose a parameter-efficient approach to fine-tune a pretrained VLM for autonomous navigation using an Imperative Learning paradigm. By optimizing against differentiable geometric cost fields rather than labeled trajectories, our model learns to generate collision-free paths exclusively from stereoscopic depth observations.
We introduce a unified end-to-end navigation pipeline for natural-language-driven robotic control. This system leverages a shared VLM backbone with task-specific Low-Rank Adaptation (LoRA) modules, effectively bridging the gap from semantic target selection to low-level trajectory planning. Our approach achieves competitive Success weighted by Path Length (SPL) in unseen environments while updating less than 1\% of the model's total parameters. Qualitative real-world experiments validate sim-to-real generalization and stable path planning without fine-tuning on real-world data. These results highlight a practical approach for deploying VLM-based agents on mobile robots, enabling high-level semantic navigation without the prohibitive requirement for large-scale, labeled trajectory data.
\end{abstract}
\section{Introduction}
Intelligent agents must seamlessly integrate perception, reasoning, and actuation to navigate unstructured and dynamic environments. In service robotics and first-response applications, agents must identify and localize context-specific entities—ranging from injured individuals to workplace personnel—to ensure socially aware and safe task execution. Besides semantic awareness, agents must also guarantee reactive, collision-free navigation, maintaining operational integrity even under significant visual domain shifts or environmental uncertainty.
\begin{figure}
    \centering
    \includegraphics[width=0.98\linewidth]{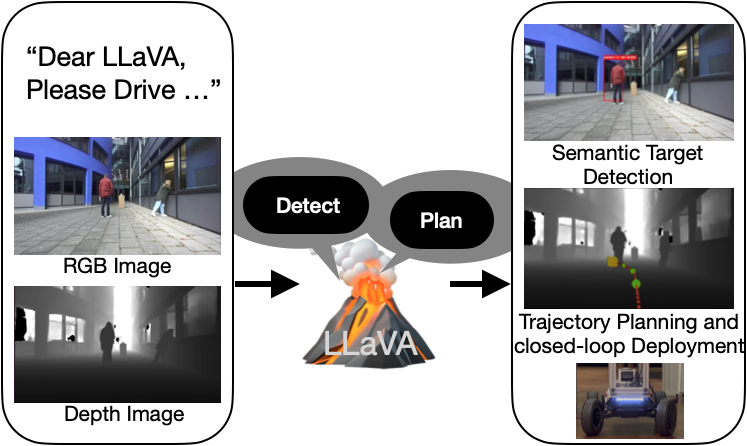}
    \caption{Semantic grounding to depth-based closed-loop trajectory planning with a VLM backbone.}
    \label{fig:eyecatcher}
\end{figure}
A primary challenge in modern robotics lies in developing versatile models capable of bridging high-level semantic reasoning with low-level geometric control. Recent advancements in Vision-Language Models (VLMs) have demonstrated remarkable proficiency in cross-modal reasoning and open-vocabulary understanding \cite{liu2024improved}. However, adapting these large-scale models for precise trajectory generation remains non-trivial. Existing VLM-based motion frameworks \cite{tang2025vlmplanner, chen2024fastnav} typically rely on supervised imitation learning, which is not only difficult to scale due to the requirement for extensive and expensive expert demonstrations, but also frequently fails to generalize to the long-tail edge cases inherent in safety-critical real-world scenarios.

To address these limitations, we propose a multi-task framework that maintains a unified VLM backbone while employing modular, parameter-efficient adaptation. By leveraging swappable Low-Rank Adaptation (LoRA) \cite{hu2022lora} modules, our approach specializes the pretrained the VLM for distinct robotic functions without the interference typical of monolithic multi-task learning. Our architecture effectively decouples the navigation pipeline: an RGB-based LoRA adapter performs open-vocabulary semantic grounding, while a depth imagery LoRA adapter, integrated with a dedicated trajectory head, generates 3D keypoints for downstream robot control. For a detailed technical description and evaluation of the object detection LoRA adapter, we refer the reader to our previous work \cite{berger2025ParameterEfficientLearningVisual}.

The planning module is trained end-to-end via the Imperative Learning paradigm \cite{yang2023iplanner}, replacing conventional imitation learning with differentiable geometric cost-field optimization. This approach enables the VLM to acquire precise navigation policies through continuous feedback without the necessity of labeled expert data. Furthermore, by utilizing depth imagery for the trajectory generation process, we minimize sensitivity to illumination and texture variations, facilitating zero-shot sim-to-real transfer.

\begin{figure}
    \centering
    \vspace{3mm}
    \includegraphics[width=0.98\linewidth]{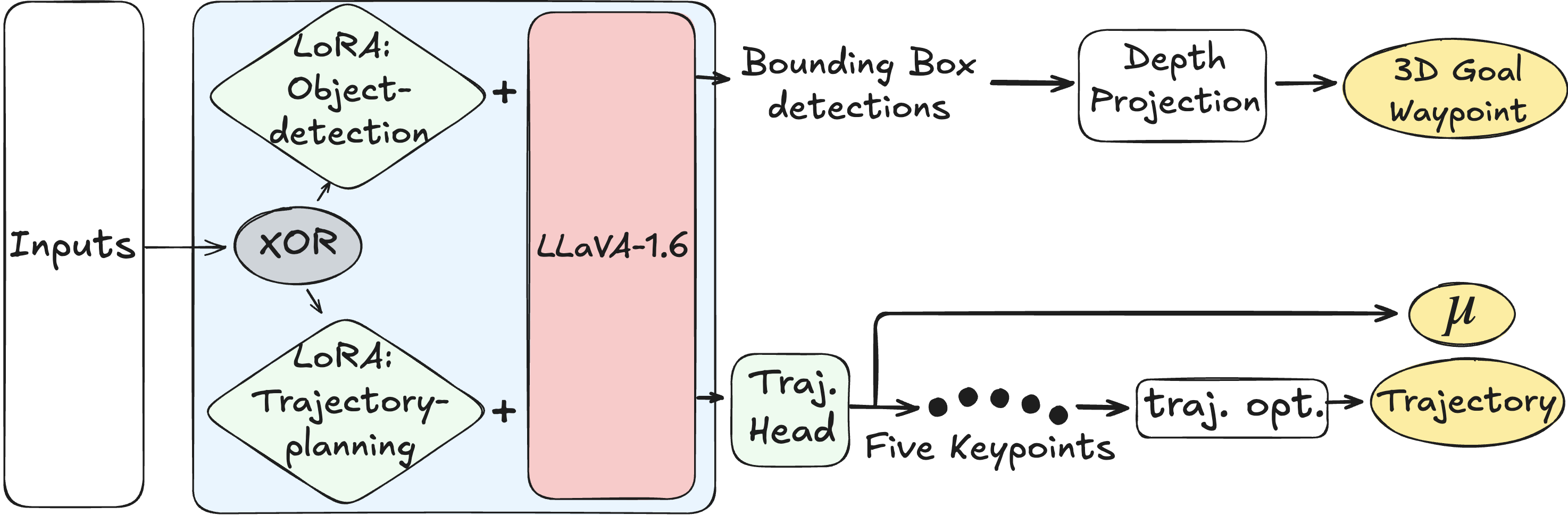}
    \caption{Overview of the modular navigation framework. A shared LLaVA-v1.6 backbone is adapted via task-specific LoRA modules. (Top) An RGB-based adapter performs open-vocabulary grounding and projects 2D detections into 3D using depth data. (Bottom) A depth-based adapter generates 3D waypoints and a collision probability $\mu$ for control.}
    \label{fig:architecture}
\end{figure}
We summarize our contributions as follows:
\begin{itemize}
    \item First, to the best of our knowledge, we are the first to integrate the Imperative Learning paradigm with parameter-efficient fine-tuning for Vision-Language Models, enabling cost-driven optimization without expert trajectory annotations.
    \item Second, we propose a modular navigation framework using a shared VLM backbone with swappable LoRA adapters, switching between RGB-based semantic grounding and depth-based trajectory planning. 
    \item Finally, we validate zero-shot sim-to-real deployment of the learned planning module on a full-scale mobile platform, showing closed-loop trajectory generation from depth observations without real-world fine-tuning.
\end{itemize}
We release model weights and training code, on GitHub\footnote{\url{https://github.com/Intelligent-Vehicles-Lab-HM/dear-llava}} to support reproducibility and further research.
\section{Related Work}
\subsection{Training VLMs}
VLMs, such as \cite{wenliang2023instructblip, Xiao_2024_florence2}, have emerged as robust architectures for tasks requiring sophisticated reasoning and contextual interpretation of visual inputs. By leveraging pretrained vision 
encoders alongside Large Language Models (LLMs), these systems 
effectively map visual observations into a high-level semantic space. 
Extensive pre-training on large-scale datasets enables VLMs to perform 
zero-shot analysis of complex scene semantics and generate coherent 
responses grounded in multimodal context. We base our work on the LLaVA-1.6 foundation model \cite{liu2024improved}, a strong state-of-the-art Vision-language baseline widely adopted in recent work.
In contrast to full-parameter fine-tuning, parameter-efficient fine-tuning strategies, such as LoRA \cite{hu2022lora}, demonstrate that optimizing a minimal subset of auxiliary parameters suffices to adapt high-capacity models for specific downstream objectives. LoRA achieves this by injecting low-rank trainable matrices into existing weight projections while maintaining the original model weights in a frozen state, thereby enabling efficient task adaptation with minimal computational and memory overhead.

\subsection{VLMs for Mobile Navigation and Path Planning}
Prior research has demonstrated the potential of VLMs for integrating visual perception and linguistic reasoning into navigation policies \cite{elnoor2025vlm, weerakoon2025behav, narasimhan2025olivianav}, yet most existing approaches remain reliant on supervised training data, costly multimodal sensor setups \cite{narasimhan2025olivianav, song2025vltgs}, or discrete intermediate representations such as cost maps \cite{weerakoon2025behav, song2025vlmsocialnav, shah2023semguesswork, sathyamoorthy2024convoi}, waypoints \cite{zhao2025imaginenav} or trajectory proposals \cite{narasimhan2025olivianav, song2025vltgs, elnoor2025vlm}.
In contrast, our method extends this line of work by coupling a pretrained VLM with the Imperative Learning paradigm \cite{yang2023iplanner}, enabling end-to-end optimization of continuous trajectories directly from depth observations without explicit trajectory supervision.  
This formulation reduces the dependency on labeled datasets and high-fidelity sensor inputs, thereby providing a parameter- and data-efficient framework for geometric reasoning in mobile robot navigation.
\section{Method} \label{Method}
\subsection{System Overview}
Our framework leverages a shared LLaVA-v1.6 backbone (CLIP \cite{radford2021learning} vision Encoder\footnote{\url{https://huggingface.co/openai/clip-vit-large-patch14-336}}, Vicuna-7B\footnote{\url{https://huggingface.co/lmsys/vicuna-7b-v1.5}} language model) to bridge semantic understanding and geometric control. As illustrated in Fig.~\ref{fig:architecture}, the system operates in two switchable phases using task-specific LoRA adapters \cite{hu2022lora}:
\begin{enumerate}
    \item \textbf{Target Grounding Phase:} The agent processes RGB input through a semantic LoRA adapter to localize high-level goals (e.g., an ``injured person''). The 2D centroid $(u,v)$ of the resulting bounding box is projected into 3D space using the synchronized depth map $\mathcal{D}$ to define a goal coordinate $g = \mathcal{D}(u,v) \cdot K^{-1}[u, v, 1]^T$, where $K$ is the intrinsic matrix.
    \item \textbf{Reactive Planning Phase:} The model hot-swaps to a depth-imagery LoRA and a trajectory head. Given the 3D goal $g$ and the current depth observation, the system generates five 3D waypoints and a collision probability $\mu$. These outputs are used for continuous, closed-loop navigation.
\end{enumerate}
\subsection{Architecture} \label{sec:architecture}
The planning module receives a depth image $D \in \mathbb{R}^{H \times W}$ replicated across three channels for compatibility with the frozen CLIP vision encoder. Following the Any-Resolution strategy \cite{liu2024improved}, $D$ is decomposed into a grid of sub-patches and a global context view, resulting in an input tensor $D_{in} \in \mathbb{R}^{5 \times 3 \times 336 \times 336}$. The goal $g$ is wrapped in a task-specific prompt and denoted by a reserved semantic token, \texttt{<goal>}, to emphasize its positional significance within the multimodal input stream.
To interpret geometric cues, we insert LoRA adapter into the language backbone's attention $(Q, K, V)$ and feed-forward blocks. Notably, the vision encoder remains entirely frozen during training, with any existing LoRA weights explicitly deactivated to ensure that the visual feature space remains untouched. The resulting multimodal hidden states $\mathbf{h} \in \mathbb{R}^{T \times d}$ are passed to the Trajectory Prediction Head, using a learnable attention-pooling layer to collapse $\mathbf{h}$ into a sequence-invariant representation $\bar{h}$. A lightweight MLP then maps $\bar{h}$ to five 3D keypoints $P \in \mathbb{R}^{5 \times 3}$ and a collision probability $\mu \in [0,1]$. Finally, the keypoints are spline-interpolated into a smooth trajectory $\tau$. 
\subsection{Imperative Objective Function}
We optimize the planning module using the imperative loss $\mathcal{L}_{imp}$ \cite{yang2023iplanner}, which replaces explicit trajectory labels with a differentiable Euclidean Signed Distance Field (ESDF) cost map. The total loss combines geometric feedback with a binary cross-entropy ``fear'' loss $\mathcal{L}_{fear}$ for the collision probability $\mu$:
\begin{equation}
    \mathcal{L} = \alpha \mathcal{C}_{obs} + \beta \mathcal{C}_{goal} + \gamma \mathcal{C}_{motion} + \lambda \mathcal{L}_{fear},
\end{equation}
where $\mathcal{C}_{obs}$ penalizes obstacle proximity, $\mathcal{C}_{goal}$ ensures target convergence, and $\mathcal{C}_{motion}$ enforces smoothness. This cost-based supervision allows the VLM to acquire navigation policies without expert human demonstrations.
\section{Experiments}
\subsection{Simulation Experiments}
We evaluate the proposed VLM-based planner against the state-of-the-art IPlanner baseline \cite{yang2023iplanner}.
Both models are trained on a set of 18 simulation environments and benchmarked on two unseen test environments: \textit{Indoors} and \textit{Forest}. 
The \textit{Indoors} environment consists of hallways and rooms containing static furniture, while the outdoor \textit{Forest} environment features a central building surrounded by tree distributions. We define 168 goal waypoints sampled from the traversable floor area across four evaluation scenarios (two for each environment). 
For a normalized comparison, both planners operate at a fixed spatial update interval of $0.15\,\text{m}$. This ensures that the evaluation metric—Success weighted by Path Length—reflects the inherent planning quality and geometric reasoning of the underlying models.
\begin{table}[ht]
    \centering
    \caption{Experimental Matrix for Real-World Validation}
    \label{tab:exp_matrix}
    \begin{tabular}{|l|l|l|}
        \hline
        \textbf{Complexity} & \textbf{Grounding (Perception)} & \textbf{Navigation (Control)} \\ \hline
        \textbf{L1} & Open-voc. grounding & Static clutter \\ \hline
        \textbf{L2} & Relational reasoning & Dynamic obs. \\ \hline
    \end{tabular}
\end{table}
\begin{figure}[t]
    \centering 
    \vspace{3mm}
    \includegraphics[width=0.98\linewidth]{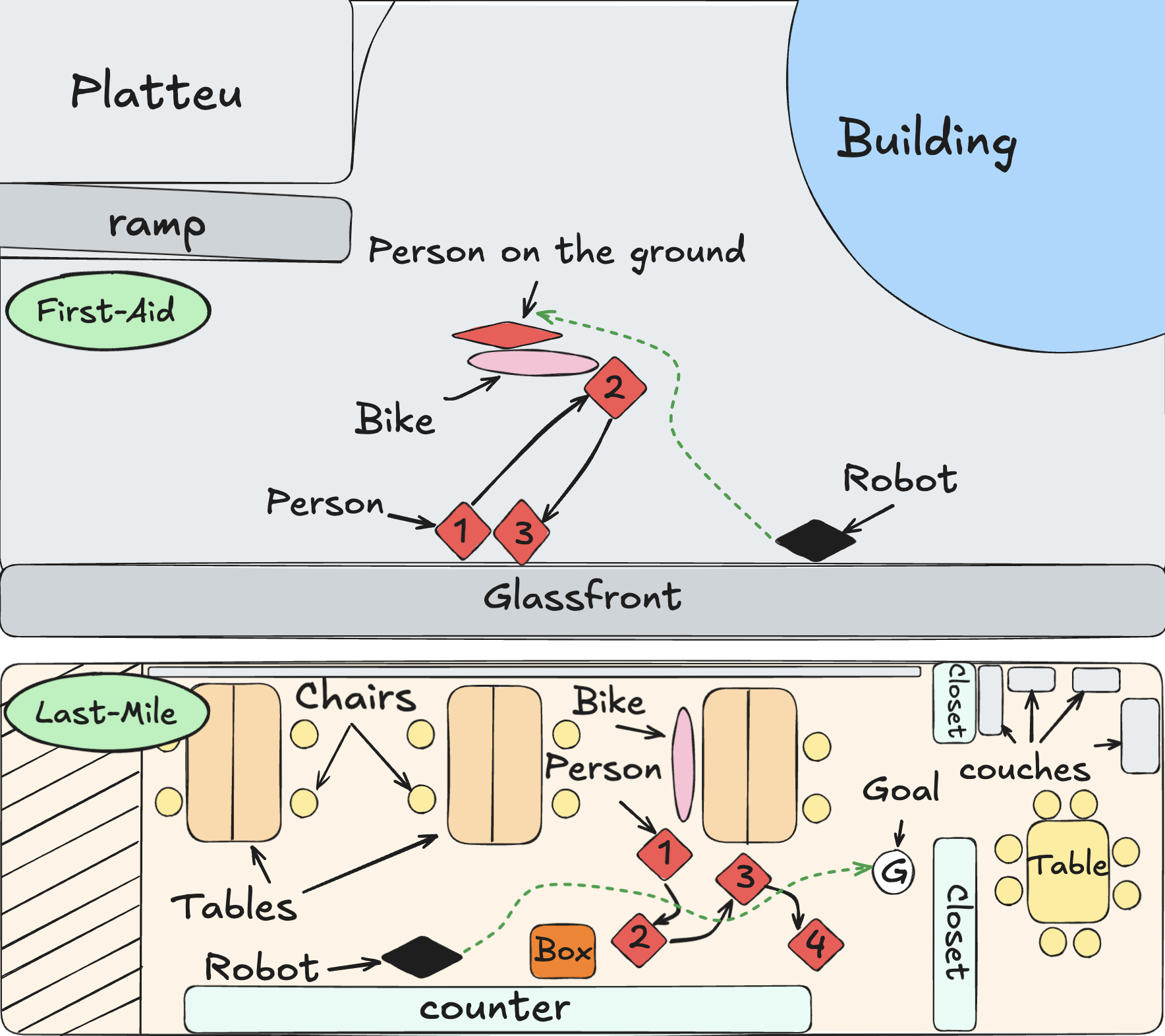}
    \caption{Example setups from the real-world experiments, showing Scenario~B, the outdoor first-aid response setup, at the top and Scenario~A, the indoor last-mile delivery setup, at the bottom.}
    \label{fig:exp_top_view}
\end{figure}
%
%
%
%
\subsection{Real-World Experiments} \label{sec:r-w-experiments}
Real-world experiments are conducted using an AgileX Hunter SE platform equipped with a ZED2i camera, an onboard Jetson Orin for perception and control, and a networked workstation hosting the user interface and agent. The robot footprint is approximately 0.87 m width by 0.82 m length with a wheelbase of 0.61 m. Evaluation is performed across two distinct real-world environments:

The indoor laboratory environment is a cluttered workspace characterized by narrow traversal corridors formed by desks, office chairs, and storage units. This environment features irregular obstacle boundaries and overhanging structures (e.g., table edges, workstations, and bins) that partly obstruct the traversable area. The approximate obstacle density, defined as the ratio of obstacle area to total workspace area,
is estimated to be $D \approx 0.45$, indicating that almost half of the floor area is occupied by static structures.

The campus environment introduces sensory challenges, including specular reflections from glass facades and complex curved geometries.
Figure \ref{fig:exp_top_view} shows an overview of our experiment environments, together with one representative experimental setup in each environment.
Our deployed pipeline follows the two-phase architecture described in Sec. \ref{sec:architecture}. 
The end-to-end control latency is \textasciitilde 1 second, measured from image acquisition to velocity command publication.
To validate the zero-shot transferability of our simulation trained model in real-world, we define an experimental matrix \ref{tab:exp_matrix} categorized by task type and complexity level. We instantiate this matrix in two service-robotics scenarios, validating the integrated pipeline under increasing semantic and geometric complexity.
We validate the zero-shot transferability of our simulation-trained model using an experimental matrix (Table~\ref{tab:exp_matrix}) instantiated in two service-robotics scenarios.

\textbf{Scenario A (Last-Mile Delivery):} This scenario evaluates the agent as a delivery platform in a cluttered office. In Requirement 1 (Level 1 Grounding), the agent localizes a semantic destination (e.g., "person at the desk") and performs spatial projection. In Requirement 2 (Level 1 Navigation), the robot traverses static clutter (furniture, boxes) to validate collision-free paths under tight geometric constraints.

\textbf{Scenario B (First-Aid Response):} This scenario tests first-responder utility in unstructured outdoor environments. Requirement 1 (Level 2 Grounding) requires the VLM to disambiguate specific targets (e.g., "injured person") using semantic context rather than generic labels. For Requirement 2 (Level 2 Navigation), the agent executes reactive trajectory updates to reach the target while adjusting to the movements of bystanders.
\subsection{Metrics} \label{sec:metrics}
For quantitative evaluations in simulation, we use the Success weighted by Path Length (SPL)~\cite{anderson2018evaluation} metric, which balances goal completion and trajectory efficiency. For each episode $i$, the SPL is defined as:
\begin{equation}
    \text{SPL}_i = S_i \cdot \frac{p_i}{\max(p_i, l_i)},
\end{equation}
where the boolean flag $S_i$ indicates success equaling $1$ if the goal is reached, and $0$ otherwise. $p_i$ is the shortest-path distance and $l_i$ the traveled distance. We average the SPL over all test episodes.

To quantify zero-shot real-world deployment over a small number of trials, we evaluate each run using three binary indicators:

\begin{itemize}
    \item \textbf{Semantic Grounding ($I$):} The specified natural-language target is correctly localized and projected into 3D goal coordinates.
    \item \textbf{Goal Reachability ($R$):} The robot terminates within $d_{thresh}=0.8$\,m of the goal.
    \item \textbf{Safety Compliance ($S$):} The run is collision-free, defined as no physical contact or wheel nudging with obstacles or humans.
\end{itemize}

We define total success as the conjunction of all applicable criteria,
\[
SR = I \land R \land S,
\]
and report raw counts per scenario to avoid overstating statistical significance.
\subsection{Training} \label{sec:training}
We train our model and reproduce the IPlanner model on the same training split using the Imperative Learning paradigm. 
Our model applies LoRA fine-tuning to the transformer projection layers ($Q$, $K$, $V$) and final fully connected layer of the LLM backbone with rank $r=16$ and scaling factor $\alpha=32$.
Only the parameters of the low-rank matrices in the language backbone and the trajectory head are updated during training, while the base model weights remain frozen. 
Our LoRA-adapted LLaVA backbone contains approximately 12.6 million trainable parameters, while the trajectory head adds about 4.9 million.  
In total, our model requires roughly 17.4 million trainable parameters.  

We use the AdamW optimizer with a learning rate between $1\times10^{-5}$ and $1\times10^{-4}$, weight decay 0.001, and early stopping with a patience of 4 epochs. The loss weights $\alpha$, $\beta$, and $\gamma$ are initialized to $0.5$, $5.0$, and $2.0$.
Training is performed on four NVIDIA A40 GPUs with a batch size of 12, using 670{,}325 image--goal pairs in total.
%
%
%
%
%
%
\section{Quantitative Results}
\begin{table}
    \centering
    \vspace{3mm}
    \caption{Comparison of our model variants against IPlanner measuring SPL across the two unseen simulation environments Indoors and Forest, each containing two navigation scenarios.}
    \label{tab:spl_results_combined}

    \setlength{\tabcolsep}{5.95pt}
    \begin{tabular}{lcccc}
        \toprule
        & \multicolumn{2}{c}{\textbf{Indoors}} & \multicolumn{2}{c}{\textbf{Forest}} \\
        \cmidrule(lr){2-3} \cmidrule(lr){4-5}
        \textbf{Model} & \textbf{Scenario~1} & \textbf{Scenario~2} & \textbf{Scenario~3} & \textbf{Scenario~4} \\
        \midrule
        IPlanner~\cite{yang2023iplanner} & 0.9748 & 0.9763 & 0.9827 & 0.9505 \\
        Ours & 0.9546 & 0.9567 & 0.9631 & 0.9223 \\
        Ours\textsuperscript{*} & 0.9250 & 0.9076 & 0.8064 & 0.6363 \\
        \bottomrule
    \end{tabular}

    \vspace{1mm}
    \begin{minipage}{\linewidth}
        \raggedright
        \footnotesize
        \textsuperscript{*}Mean-pooling variant.
    \end{minipage}
\end{table}
As shown in Table \ref{tab:spl_results_combined}, our VLM-based planner achieves SPL scores comparable to the specialized IPlanner baseline across all simulation environments (over 97\% of IPlanners SPL performance). This results demonstrates that a large-scale multimodal backbone can be adapted for low-level geometric control without a large performance mitigation in precision relative to task-specific models. While IPlanner is limited to trajectory generation, our model maintains the inherent capacity for disparate downstream tasks, like open-vocabulary target identification.
\section{Qualitative Results} \label{Results}
\begin{figure*}
    \centering
    \vspace{3mm}
    \newcommand{\imgw}{\linewidth}
    \newcommand{\imgh}{0.18\textheight}

    \begin{subfigure}[t]{0.19\textwidth}
        \centering
        \includegraphics[width=\imgw,height=\imgh,keepaspectratio]{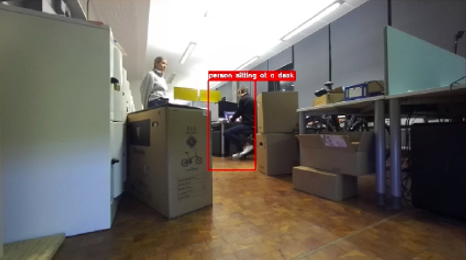}\\[0.4em]
        \includegraphics[width=\imgw,height=\imgh,keepaspectratio]{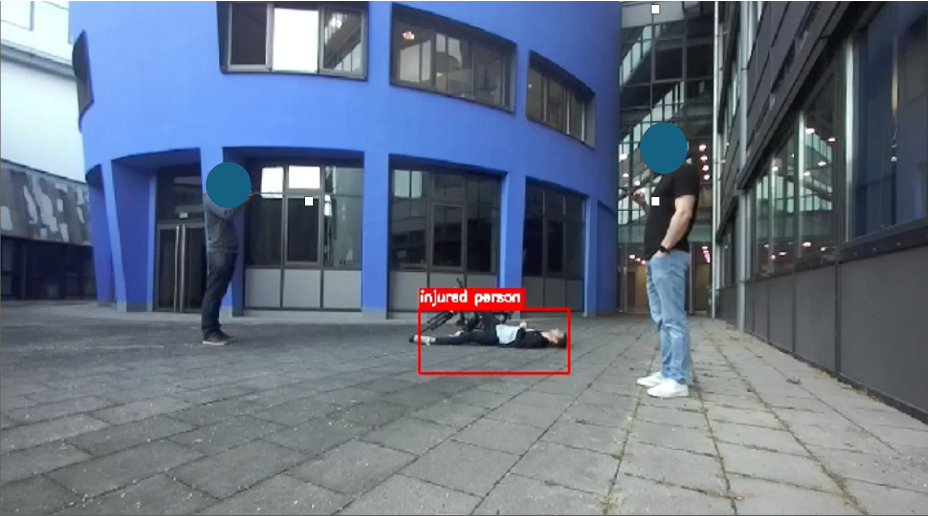}
    \end{subfigure}\hfill
    \begin{subfigure}[t]{0.19\textwidth}
        \centering
        \includegraphics[width=\imgw,height=\imgh,keepaspectratio]{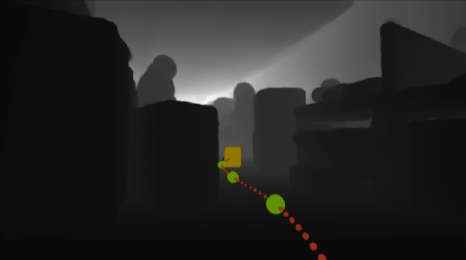}\\[0.4em]
        \includegraphics[width=\imgw,height=\imgh,keepaspectratio]{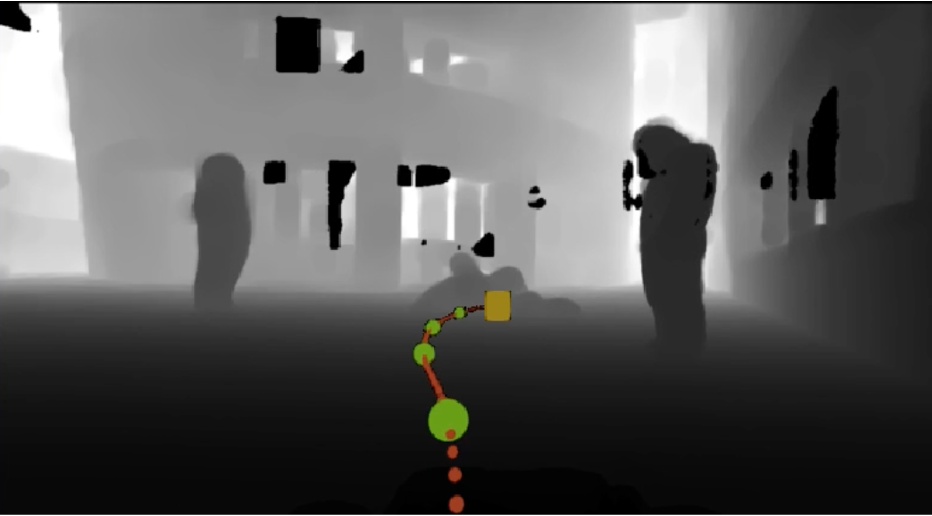}
    \end{subfigure}\hfill
    \begin{subfigure}[t]{0.19\textwidth}
        \centering
        \includegraphics[width=\imgw,height=\imgh,keepaspectratio]{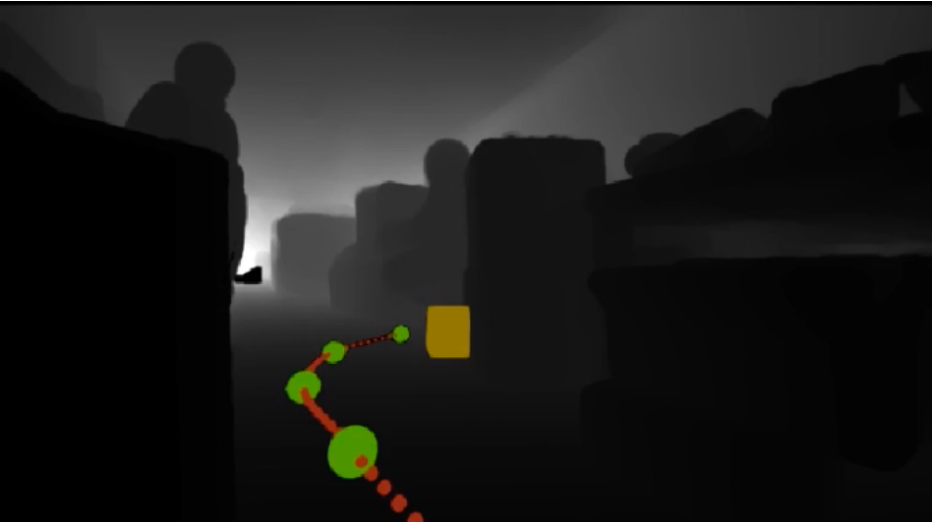}\\[0.4em]
        \includegraphics[width=\imgw,height=\imgh,keepaspectratio]{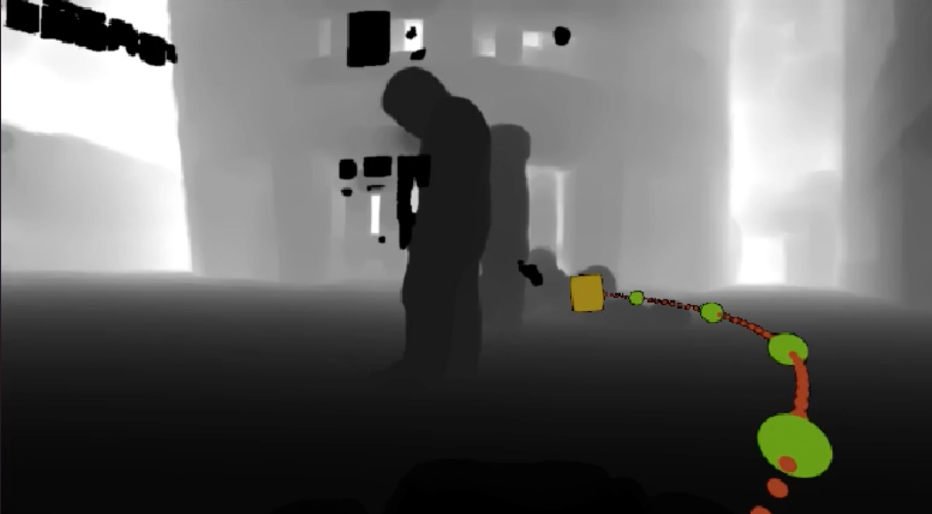}
    \end{subfigure}\hfill
    \begin{subfigure}[t]{0.19\textwidth}
        \centering
        \includegraphics[width=\imgw,height=\imgh,keepaspectratio]{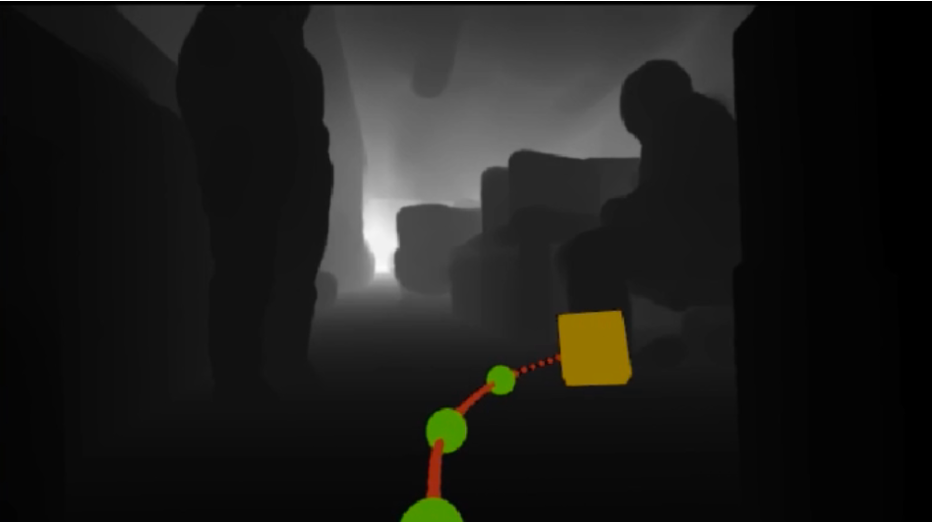}\\[0.4em]
        \includegraphics[width=\imgw,height=\imgh,keepaspectratio]{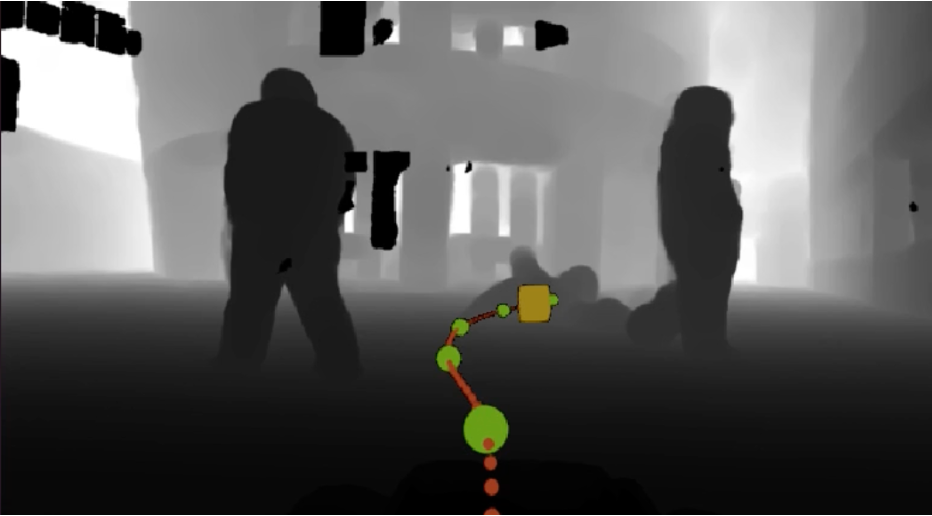}
    \end{subfigure}\hfill
    \begin{subfigure}[t]{0.19\textwidth}
        \centering
        \includegraphics[width=\imgw,height=\imgh,keepaspectratio]{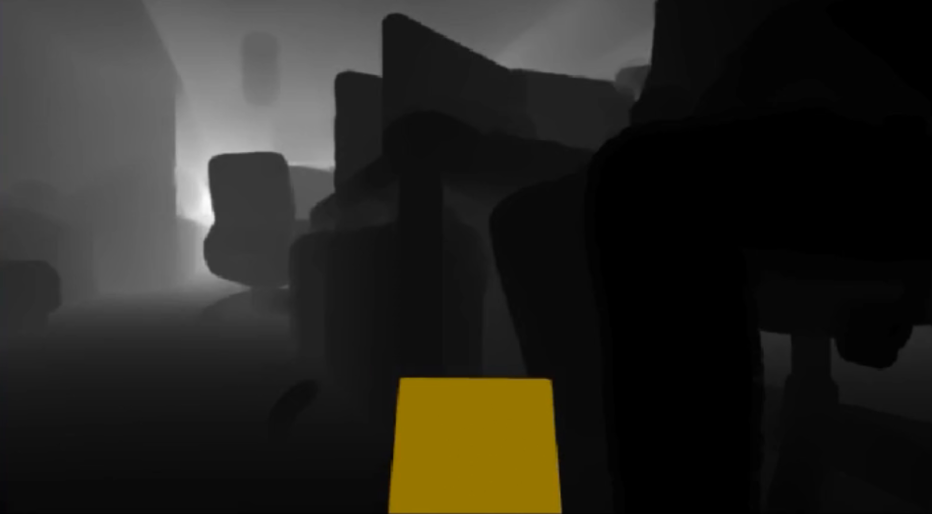}\\[0.4em]
        \includegraphics[width=\imgw,height=\imgh,keepaspectratio]{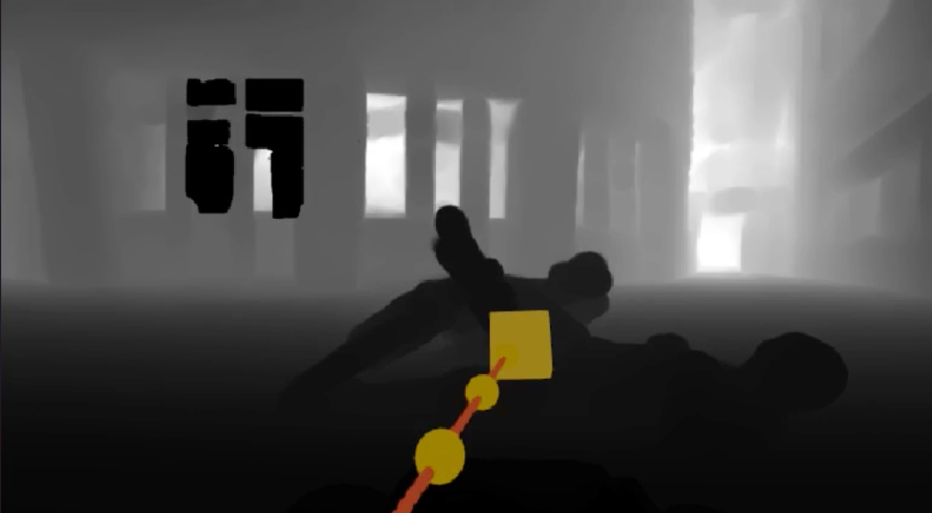}
    \end{subfigure}

    \caption{
    Qualitative execution traces for real-world validation. 
    (Top) Delivery scenario: The agent performs open-vocabulary grounding for the query ``person sitting at the desk'' and executes a trajectory to the target. 
    (Bottom) First-aid scenario: The agent identifies an ``injured person'' and successfully terminates the approach before physical contact, by classifying the trajectory as invalid based on the predicted collision probability $\mu$.
    }
    \label{fig:real_world_exp}
\end{figure*}

During simulation, the proposed planner consistently maintains larger obstacle margins than IPlanner. While this behavior results in slightly longer paths and reduced SPL, it produces fewer near-boundary trajectories. As both models were trained under identical conditions, the behavioral difference appears to stem from the influence of the pretrained backbone. A deeper analysis of this effect is left for future work.
\begin{table}[ht]
    \centering
    \caption{Summary of Real-World Experiments. $I$: Semantic Grounding, $R$: Goal Reachability, $S$: Safety Compliance, $SR$: Success Rate ($I \land R \land S$).}
    \label{tab:realworld_summary}
    \begin{tabular}{@{}lccccc@{}}
        \toprule
        \textbf{Use Case} & \textbf{Trials} & \textbf{$I$} & \textbf{$R$} & \textbf{$S$} & \textbf{$SR$} \\ \midrule
        Last-Mile Delivery & 20 & 18 & 16 & 12 & 12 \\
        First-Aid Response & 15 & 13 & 11 & 13 & 11 \\ \midrule
        \textbf{Total / Avg.} & \textbf{35} & \textbf{31} & \textbf{27} & \textbf{25} & \textbf{65.7\%} \\ \bottomrule
    \end{tabular}
\end{table}
We deploy the simulation-trained model in real-world environments to asses its sim-to-real robustness. Operating without real-world fine-tuning, the system is assessed across the scenarios defined in Section~\ref{sec:r-w-experiments}. Figure~\ref{fig:real_world_exp} provides a qualitative execution trace of these trials, illustrating the transition from initial open-vocabulary target grounding to reactive trajectory updates generated from egocentric depth observations.
We observe that, despite the slow update rate, the generated trajectories remain adaptive in the presence of dynamic obstacles. 
In cases where the predicted trajectory intersected obstacles in the depth observation, the associated collision probability $\mu$ increased and led to termination when exceeding the threshold of 0.5. 
Across all trials, no body-level collisions occurred along the predicted center-of-mass trajectories. Minor wheel-level contacts were observed in narrow passages during execution. These events were attributable to the downstream path-following controller operating on a simplified geometric abstraction and not explicitly accounting for the full footprint and kinematic constraints of the deployed platform. The planned trajectories themselves remained geometrically feasible within the planner’s representation.
In addition, bounding boxes for distant entities occasionally overestimated spatial extent, resulting in slightly imprecise 3D goal projections.
\begin{figure}[!t]
    \centering
    \begin{subfigure}[t]{0.48\linewidth}
        \centering
        \vspace{3mm}
        \includegraphics[width=\linewidth]{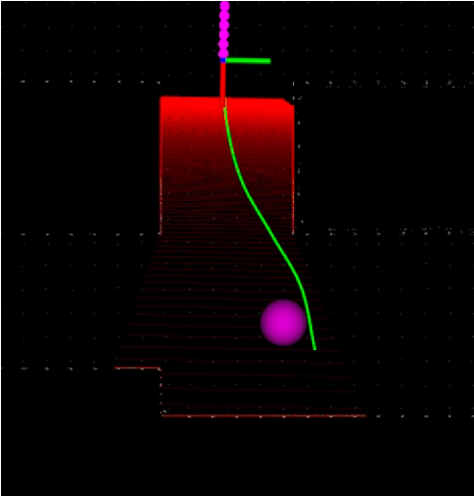}
        \caption{Mean-pooling}
        \label{fig:abl_first_model}
    \end{subfigure}\hfill
    \begin{subfigure}[t]{0.48\linewidth}
        \centering
        \vspace{3mm}
        \includegraphics[width=\linewidth]{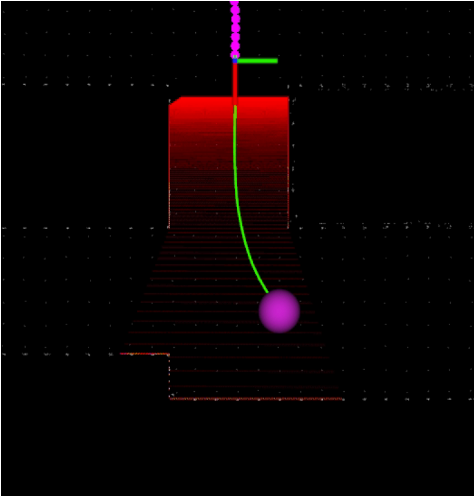}
        \caption{Attention-pooling}
        \label{fig:abl_fixed_goal}
    \end{subfigure}
    \caption{Trajectory predictions obtained with our trajectory head applying (a) mean pooling versus (b) attention pooling from the last hidden states.}
    \label{fig:Att_pool_improvements}
\end{figure}
\section{Ablation Studies} \label{sec:ablations}
In this section, we present two ablation studies: We (1) explore the effects of integrating a mean-pooling versus an attention-pooling layer in the trajectory head and (2) evaluate model performance utilizing depth inputs against RGB images for perception.
\subsection{Trajectory Head}
To evaluate the impact of the attention-pooling mechanism described in Section~\ref{sec:architecture}, we implement a mean-pooling variant for comparison. Both models share the same backbone and training configuration, differing only in the pooling strategy used to aggregate the multimodal hidden states across the sequence dimension $T$.
As shown in Table \ref{tab:spl_results_combined}, the attention-aware variant achieves a 2.4\% higher SPL on average. This performance gain is further supported by the qualitative comparison in Fig. \ref{fig:Att_pool_improvements}. While the mean-pooling baseline (Fig. \ref{fig:abl_first_model}) exhibits trajectory drift, the attention-pooling head (Fig. \ref{fig:abl_fixed_goal}) demonstrates superior goal fixation, producing smoother paths that converge more accurately on the target center.
\subsection{Depth vs. RGB}
%
We evaluate the impact of depth inputs on domain generalization by comparing a depth-based model and an RGB-based model trained on an identical subset of 12 simulation environments containing synchronized RGB imagery. The remaining six environments in the full dataset did not provide RGB data and were therefore excluded from this ablation. As shown in Table \ref{tab:rgb}, the depth-based model consistently outperforms the RGB variant across all scenarios. In the Forest environment, RGB fails to navigate Scenario 3 entirely and is surpassed by the depth-based model by 40.1\% in Scenario 4. 
\begin{figure}[ht]
    \centering
    \begin{subfigure}[t]{0.48\linewidth}
        \centering
        \vspace{3mm}
        \includegraphics[width=\linewidth]{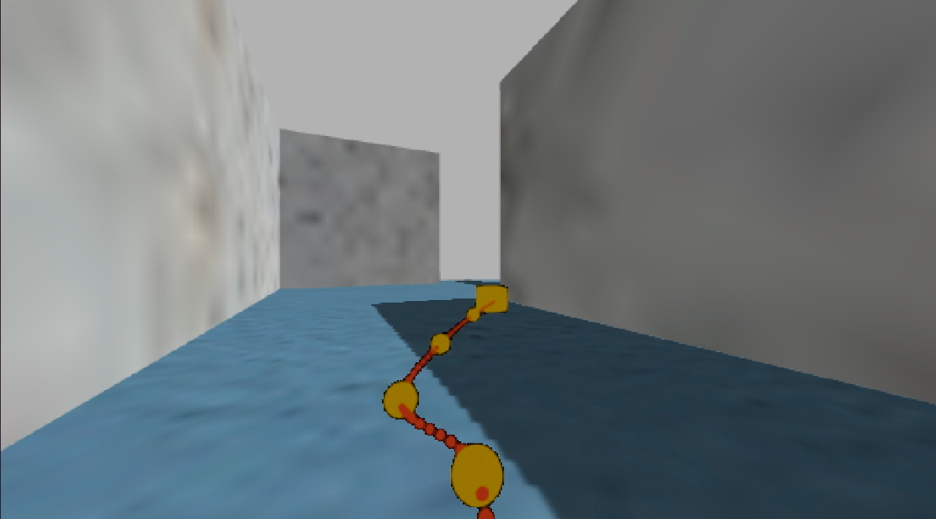}
        \label{fig:rgb_failed}
    \end{subfigure}\hfill
    \begin{subfigure}[t]{0.48\linewidth}
        \centering
        \vspace{3mm}
        \includegraphics[width=\linewidth]{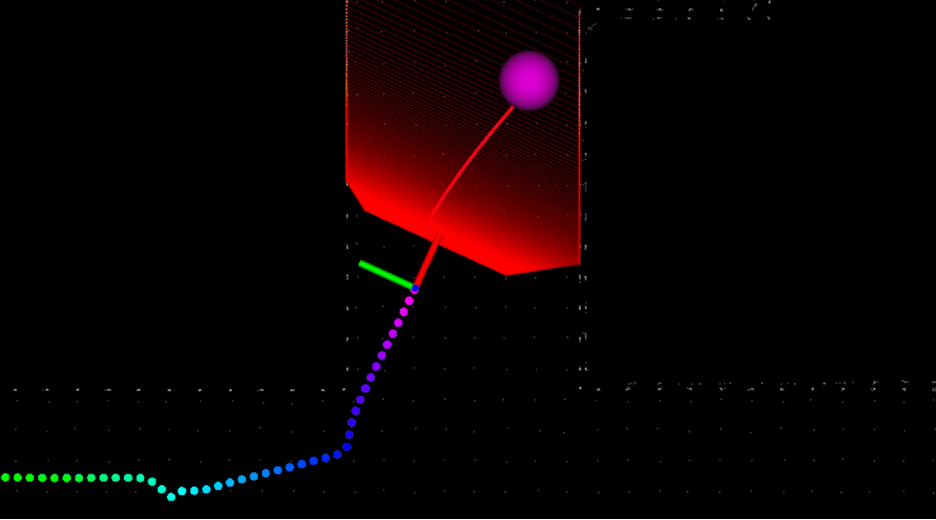}
        \label{fig:RGB_view}
    \end{subfigure}
    \caption{Example of misclassification of a trajectory as invalid in traversable region in scenario 1 of the Indoor environment (left) and the corresponding BEV view of the scene (right).}
    \label{fig:rgb_fear_paths}
\end{figure}
The RGB-based model's underperformance stems from predicting high collision fear in traversable areas, often mistaking shadows for obstacles (Fig. \ref{fig:rgb_fear_paths}). This issue is absent with depth inputs, demonstrating that depth-based perception provides higher generalization capabilities to domain shift, achieving an SPL rate of 89\% or higher in unseen environments.
%
%
%
%
\section{Discussion}
Our findings indicate that the Imperative Learning paradigm with geometric cost supervision enables parameter-efficient adaptation of pretrained Vision-Language Models for trajectory generation. Without labeled expert demonstrations, our model achieves over 97\% of IPlanner’s SPL performance while updating less than 1\% of the backbone parameters and keeping the vision encoder completely frozen. These results demonstrate that large multimodal representations can be specialized for low-level geometric control using lightweight adaptation modules. 
\begin{table}[ht]
    \centering
    \caption{Comparison of RGB and Depth Perception (SPL)}
    \label{tab:rgb}
    \vspace{3pt}
    \setlength{\tabcolsep}{4pt}
    \begin{tabularx}{\linewidth}{l *{4}{>{\centering\arraybackslash}X}}
        \toprule
        & \multicolumn{2}{c}{\textbf{Indoors}} & \multicolumn{2}{c}{\textbf{Forest}} \\
        \cmidrule(lr){2-3} \cmidrule(lr){4-5}
        \textbf{Model} & \textbf{Scenario 1} & \textbf{Scenario 2} & \textbf{Scenario 3} & \textbf{Scenario 4} \\
        \midrule
        Ours (RGB)   & 0.8565 & 0.8816 & 0.0    & 0.6362 \\
        Ours (Depth) & 0.9367 & 0.9226 & 0.9353 & 0.8935 \\
        \bottomrule
    \end{tabularx}
\end{table}

Despite these promising results, several limitations emerged during real-world deployment. First, the system relies on the ZED2i integrated odometry, which exhibited drift and occasional discontinuities in the global frame. These inconsistencies affected long-range waypoint stability. 
Second, the current end-to-end inference latency of approximately 1 second necessitates conservative velocity profiles and limits reactive control in highly dynamic environments. While specialized planners prioritize lower latency, utilizing a VLM establishes a foundation for context-aware navigation. Validating VLM efficacy in low-level geometric control serves as a critical prerequisite for future architectures that integrate spatial feasibility with high-level semantic, historical, and socially compliant reasoning.

Finally, the current pipeline operates under a simplified geometric abstraction and does not explicitly model the full kinematic and footprint constraints of the deployed platform. As observed in narrow passages, this abstraction can lead to embodiment-related discrepancies during execution.
\section{Conclusion}
In this work, we presented a depth-aware Vision-Language Agent for closed-loop robotic control that connects semantic target grounding with geometric trajectory generation. By integrating the Imperative Learning paradigm with parameter-efficient fine-tuning (LoRA), we demonstrated that a pretrained VLM can be adapted for trajectory prediction using differentiable geometric cost fields instead of labeled expert demonstrations.

Across unseen simulation environments, our model achieves performance comparable to a task-specific planner while updating less than 1\% of the backbone parameters. Real-world deployment without additional fine-tuning further validates the feasibility of zero-shot sim-to-real transfer under practical sensing.

Future work will focus on reducing inference latency by exploring more recent, parameter efficient VLM backbones, incorporating kinematic-constrained control, and improving state estimation robustness to enhance deployment stability in more dynamic and large-scale environments.

\section{Acknowledgement}
The research leading to these results is funded by the German Federal Ministry for Economic Affairs and Energy within the project “NXT GEN AI METHODS – Generative Methoden für Perzeption, Prädiktion und Planung". The authors would like to thank the consortium for the successful cooperation. The authors gratefully acknowledge the scientific support and HPC resources provided by the Erlangen National High Performance Computing Center (NHR@FAU) of the Friedrich-Alexander-Universität Erlangen-Nürnberg (FAU). The hardware is partially funded by the German Research Foundation (DFG).

\bibliographystyle{IEEEtran}
\bibliography{refs}
\end{document}